\documentclass[letterpaper, 10pt, conference]{ieeeconf}  % Comment this line out if you need a4paper

\IEEEoverridecommandlockouts                              % This command is only needed if 
\usepackage{graphicx}
\usepackage{amsmath}
\usepackage{amsthm}
\usepackage{amssymb}
\usepackage{booktabs}
\usepackage[utf8]{inputenc}
\usepackage{tikz}
\usepackage{pgfplotstable}
\usepackage{pgfplots}
\usepackage{subcaption}
\usepackage{caption}
\usepackage[font=footnotesize]{caption}
\usepgfplotslibrary{statistics}
\usetikzlibrary[calc]
\usetikzlibrary[pgfplots.groupplots]
\usepackage{enumerate}
\usepackage{rotating}
\usepackage{todonotes}
\usepackage{arydshln}
\usepackage{cite}
\usepackage{multirow}
\usepackage{comment}
\usepackage[compact]{titlesec}
\titlespacing{\subsubsection}{0pt}{*0}{*0}
\usepackage{wrapfig}
\usepackage{adjustbox}
\usepackage[linesnumbered,ruled,vlined]{algorithm2e}
\SetKwComment{Comment}{$\triangleright$\ }{}

\SetCommentSty{mycommfont}

\newtheoremstyle{mystyle}%                % Name
  {}%                                     % Space above
  {}%                                     % Space below
  {\itshape}%                                     % Body font
  {}%                                     % Indent amount
  {\bfseries}%                            % Theorem head font
  {.}%                                    % Punctuation after theorem head
  { }%                                    % Space after theorem head, ' ', or \newline
  {\thmname{#1}\thmnumber{ #2}\thmnote{ (#3)}}%                                     % Theorem head spec (can be left empty, meaning `normal')

\usepackage{hyperref}
\usepackage{xcolor}
\definecolor{medium-blue}{rgb}{0,0,1}
\hypersetup{colorlinks, urlcolor={medium-blue}}
\theoremstyle{mystyle}
\newtheorem{theorem}{Conjecture}
\newtheorem{proposition}{Definition}

\title{\LARGE \bf GMCR: Graph-based Maximum Consensus Estimation \\for Point Cloud Registration}

\author{Michael Gentner, Prajval Kumar Murali and Mohsen Kaboli% <-this % stops a space
\thanks{M. Gentner, P.K. Murali, and M. Kaboli are with the BMW Group, Munich, Germany. 
e-mail: name.surname@bmwgroup.com}%
\thanks{M. Gentner is with the Technical University of Munich, Germany}%
\thanks{P.K. Murali is with the University of Glasgow, Scotland}%
\thanks{M. Kaboli is with the Donders Institute for Brain and Cognition, Radboud University, Netherlands }%
}

\begin{document}
\maketitle
\thispagestyle{empty}
\pagestyle{empty}

%%%%%%%%%%%%%%%%%%%%%%%%%%%%%%%%%%%%%%%%%%%%%%%%%%%%%%%%%%%%%%%%%%%%%%%%%%%%%%%%
\begin{abstract}

% Intro sentence
Point cloud registration is a fundamental and challenging problem for autonomous robots interacting in unstructured environments for applications such as object pose estimation, simultaneous localization and mapping, robot-sensor calibration, and so on. In global correspondence-based point cloud registration, data association is a highly brittle task and commonly produces high amounts of outliers. Failure to reject outliers can lead to errors propagating to downstream perception tasks. Maximum Consensus (MC) is a widely used technique for robust estimation, which is however known to be NP-hard. Exact methods struggle to scale to realistic problem instances, whereas high outlier rates are challenging for approximate methods. To this end, we propose Graph-based Maximum Consensus Registration (GMCR), which is highly robust to outliers and scales to realistic problem instances. We propose novel consensus functions to map the decoupled MC-objective to the graph domain, wherein we find a tight approximation to the maximum consensus set as the maximum clique. The final pose estimate is given in closed-form. We extensively evaluated our proposed \textit{GMCR} on a synthetic registration benchmark, robotic object localization task, and additionally on a scan matching benchmark. 
Our proposed method shows high accuracy and time efficiency compared to other state-of-the-art MC methods and compares favorably to other robust registration methods.

\end{abstract}

%%%%%%%%%%%%%%%%%%%%%%%%%%%%%%%%%%%%%%%%%%%%%%%%%%%%%%%%%%%%%%%%%%%%%%%%%%%%%%%%
\section{Introduction}\label{sec:introduction}
%% Motivation
Robust perception is critical for the increased autonomy of a robot equipped with various sensing modalities in unstructured environments. It deals with the estimation of unknown quantities (for instance, object pose, self-localization, and environment mapping) from potentially noisy or partial sensor measurements. Algorithms for robust perception need to handle high outlier rates, strong non-linearities, and high uncertainty levels~\cite{bosse2016robust}. Point cloud registration is one of the most studied robust perception tasks and finds its applications in object pose estimation~\cite{murali2022active}, simultaneous localization and mapping~\cite{pomerleau2013comparing}, 3D reconstruction~\cite{delmerico2018comparison}, scan matching~\cite{magnusson2015beyond} and medical imaging~\cite{audette2000algorithmic}. It is not limited to vision or range-based sensors but is also used for tactile sensing as shown by recent works~\cite{li2020review, murali2021active, murali2022intelligent, kaboli2019tactile, kaboli2018active, kaboli2017tactile, kaboli2018robust, liu2022neuro}.   

\begin{figure}[t!]
    \centering
    \includegraphics[width=\columnwidth]{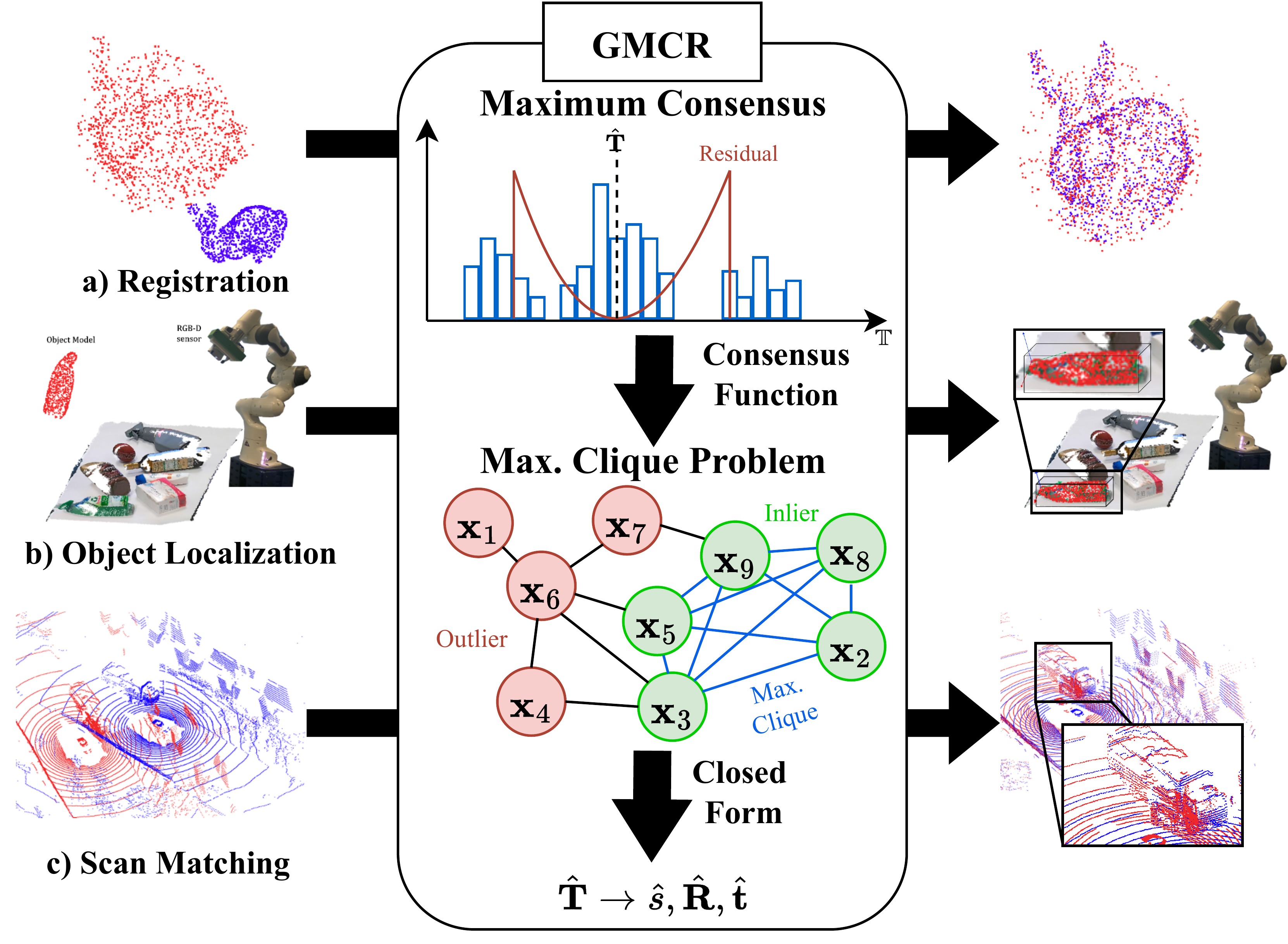}
    \caption{Our proposed robust registration approach GMCR maps the Maximum consensus objective which is \textit{NP-hard} to a Graph Clique problem, where the solution can be found in closed form from the maximum clique. We evaluated GMCR on the (a) Stanford scanning repository benchmark, (b) a real robot object localization setup, and (c) a real-world LiDAR scan benchmark.}
    \label{fig:fig1}
\end{figure}

Rigid point cloud registration is defined by finding the optimal similarity transformation aligning two 3D point sets $\mathcal{A}$ and $\mathcal{B}$. Global registration is often approached by subsequent point-to-point correspondence estimation and fitting. Since data association in 3D is highly error prone~\cite{Bustos2018, Parra2019a}, methods must reliably reject outliers. Furthermore, point-wise features are often computed with local descriptors based on point statistics. This can result in incorrect correspondence generation and even spatially correlated outliers resulting from ambiguous regions. The challenge of structured noise has also been studied in 2D and 3D~\cite{guerrero2018pcpnet, jalobeanu2014unknown, 4458156}. A popular robust objective is Maximum Consensus (MC)~\cite{10.5555/993884}, which finds inliers as the largest set of correspondences that can be aligned by one common estimate (see figure \ref{fig:fig1}). Note that MC, as other robust objectives such as Truncated-Least-Squares (TLS) and Least-Trimmed-Squares (LTS), still can return an arbitrarily wrong estimate depending on the input~\cite{carlone2022estimation}. However, there clearly exist settings in which some objectives perform better than others. Maximum Consensus in its exact form is \textit{NP-hard}~\cite{Chin}. Stochastic techniques, such as RANSAC~\cite{fischler1981random} have been proposed, which randomly sample correspondence subsets and compute the consensus set. Many extensions exists, which increase speed~\cite{10.1007/3-540-47969-4_6, 1467271}, improve accuracy~\cite{torr2002bayesian} and increase robustness~\cite{konouchine2005amlesac, sun2021ransic}. The convergence speed of RANSAC however scales exponentially with the outlier rate~\cite{Bustos2018} and the result can be arbitrarily wrong upon termination. Many globally optimal methods are based on searching through the whole solution space using \textit{Branch-and-Bound (BnB)}~\cite{bazin2012globally, zheng2011deterministically, olsson2008branch}. 
This has been extended with MIP~\cite{li2009consensus}, tree search~\cite{chin2015efficient}, and new bounding functions~\cite{parra2014fast} to speed up computation. In the general case, BnB has an exponential runtime in problem size and outlier rate~\cite{Bustos2018}, which makes it hardly scalable for large problem sizes. Recently Speciale \textit{et al.}~\cite{Speciale2017} formulate MC as a Mixed-Integer-Program (MIP) and show that introducing Linear-Matrix-Inequality Constraints speeds up computation substantially. However, the computation time still increases with the outlier rate and problem size. Robust sums-of-squares have also been used, but are challenging to solve globally optimal for realistic problems~\cite{Yang2020}. Recently, Yang \textit{et al.}~\cite{Yang2021} proposed TEASER++ that uses a Truncated-Least-Squares (TLS) cost and leverages Graduated-non-Convexity (GNC) to solve the rotation objective. GNC however also struggles with very high outlier rates~\cite{Bustos2018} and limited performance~\cite{carlone2022estimation}. Fast-Global-Registration (FGR)~\cite{Zhou2016} also uses a robust least-squares cost and reformulates it with Black-Rangarajan duality~\cite{Black1996} to solve it efficiently. Simultaneous Pose and Correspondence (SPC) methods are the second popular paradigm in registration with the pioneering work ICP~\cite{Besl1992}. Numerous robust extensions have been proposed~\cite{Granger2002, Kaneko2003, Chetverikov2005, Maier-Hein2012}. Segal \textit{et al.} combine point-to-point and point-to-plane ICP in Generalized ICP (GICP)~\cite{segal2009generalized}. SPC methods are usually very fast and accurate but are highly susceptible to local minima~\cite{pomerleau2013comparing}. Recently, \textit{Murali et al.}~\cite{murali2021active, murali2022active, murali2022empirical} proposed a translation-invariant quaternion filter (TIQF) which is a filtering-based approach for registration and focuses on extremely sparse point clouds. Global SPC methods such as Go-ICP~\cite{Yang2016} were proposed. Most of the global methods are however based on BnB~\cite{hartley2009global, breuel2003implementation}, which then again results in computationally expensive optimization. Recently, deep learning-based methods for registration have emerged~\cite{guo2020deep, Aoki2019, Wang2019a, Wang2019b}. Deep learning-based methods, although showing great promise, exhibit not well-understood behavior when confronted with noise and outliers~\cite{wicker2019robustness}. Methods to reject outliers without solving the registration task have been proposed. Bustos \textit{et al.} introduce GORE~\cite{Bustos2018}, which provably only removes outliers, which are not in the inlier set of the corresponding maximum consensus objective. Consistency graph-based methods encode consistency between pairs of correspondences using invariants~\cite{Enqvist2009, 8460217, leordeanu2005spectral} in a graph. Outliers are then rejected by finding the inlier set as the maximum clique. More efficient search algorithms~\cite{Parra2019a} and relaxations to the maximum clique problem~\cite{Lusk2021} have been proposed. Shi \textit{et al.}~\cite{Shi2020} generalize the concept of invariants and efficiently find the inlier set as the maximum k-core. Consistency graph-based methods however require the definition of an invariant (a quantity invariant under transformation~\cite{Lusk2021}).  \\
\textbf{GMCR (Ours) vs. Consistency Graph-based methods: }
Consistency graph-based methods are used for fast outlier rejection using invariants~\cite{Lusk2021, Shi2020, Parra2019a, Enqvist2009, 8460217}. 
\begin{comment}
    PointNetLK~\cite{Aoki2019} minimizes the feature reprojection error in 3D and Deep Closest Point (DCP)~\cite{Wang2019a} computes the estimate based on correspondences extracted with a Dynamic Graph CNN~\cite{Wang2019b}. 
% Yang \textit{et al.}~\cite{yang2021self} combine robust fitting and deep feature extraction in a self-supervised framework.
Deep learning-based methods, although showing great promise, exhibit not well-understood behavior when confronted with noise and outliers~\cite{wicker2019robustness}.
\end{comment}
\begin{comment}
\begin{wrapfigure}[11]{r}{0.6\linewidth}
  \centering
  \vspace{-15pt}
    \includegraphics[width=\linewidth]{ICRA_2023/figures/1_introduction/cons_graph.pdf}
    \vspace{-10pt}
  \caption{Example of Scale Consistency Graphs vs. GMCR Scale Graphs}
  \label{fig:graph_consistency}
\end{wrapfigure}
\end{comment}
A graph is built where nodes represent associations, edges encode consistency, and the inlier set is given by the maximum clique. An invariant can thus, by definition~\cite{Lusk2021}, not find the inlier set for an estimation problem such as MC. On the other hand, our graph is built such that the maximum clique represents the inlier set of the maximum consensus estimate. Nodes represent measurements stemming from pairs of associations and edges encode consensus. \\

%% Contributions
\textbf{Contributions:} \\
Robust point cloud registration algorithms still struggle with high outlier rates, scalability to a large number of associations, and various types of noise in the measurements. Furthermore, the NP-hard nature of the MC-objective requires approximate approaches to scale to realistic problem instances.\\
%Furthermore, sampling-based methods cannot guarantee the quality of the solution. (Prajval: not sure if we can provide guarentees).
To this end, we propose the following contributions:
\begin{enumerate}[{(I)}]
    \setlength\itemsep{0em}
    \item We propose a graph-based method to solve the maximum consensus objective for registration (GMCR). We build graphs representing the consensus structure within the data and find the maximum consensus set as the maximum clique. 
    \item We propose novel consensus functions to extract the graph structures. Our proposed framework results in decreasing runtime with increasing outliers while being robust to various types of outliers. 
    \item We perform extensive experiments on standard benchmarks, robot experiments, and scan-matching datasets for point cloud registration and demonstrate that GMCR compares favorably with state-of-the-art methods in terms of accuracy and timing.
\end{enumerate}

\section{Methodology}\label{sec:method}
\begin{figure*}[t!]
     \centering
     \begin{subfigure}[b]{0.27\textwidth}
         \centering
         \includegraphics[width=\textwidth]{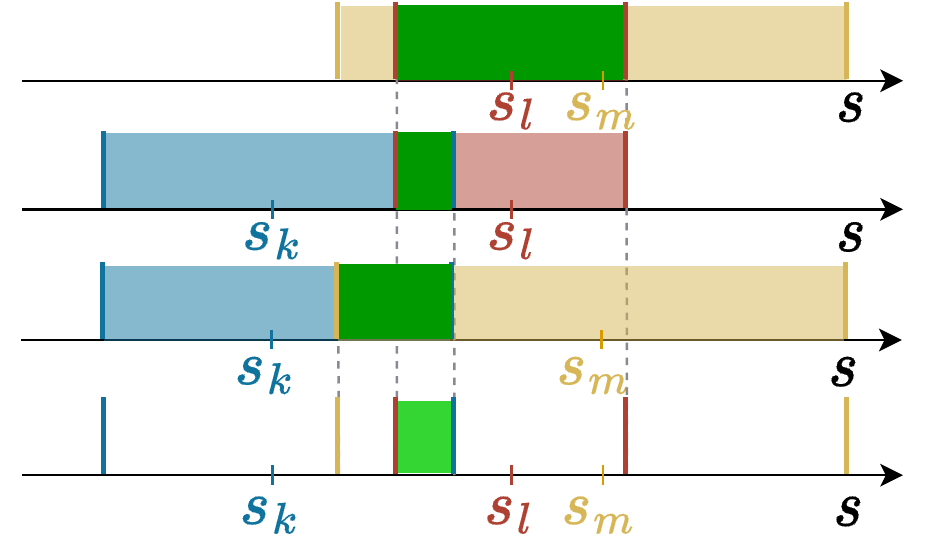}
         \caption{Intervals around scale measurements $s_l$, $s_m$, and $s_k$. $\mathbb{I}_s = 1$ for all three pairs $(s_l, s_m)$, $(s_l, s_k)$ and $(s_m, s_k)$.}
         \label{fig:consensus_func:scale}
     \end{subfigure}
     \hfill
     \begin{subfigure}[b]{0.27\textwidth}
         \centering
         \includegraphics[width=.8\textwidth]{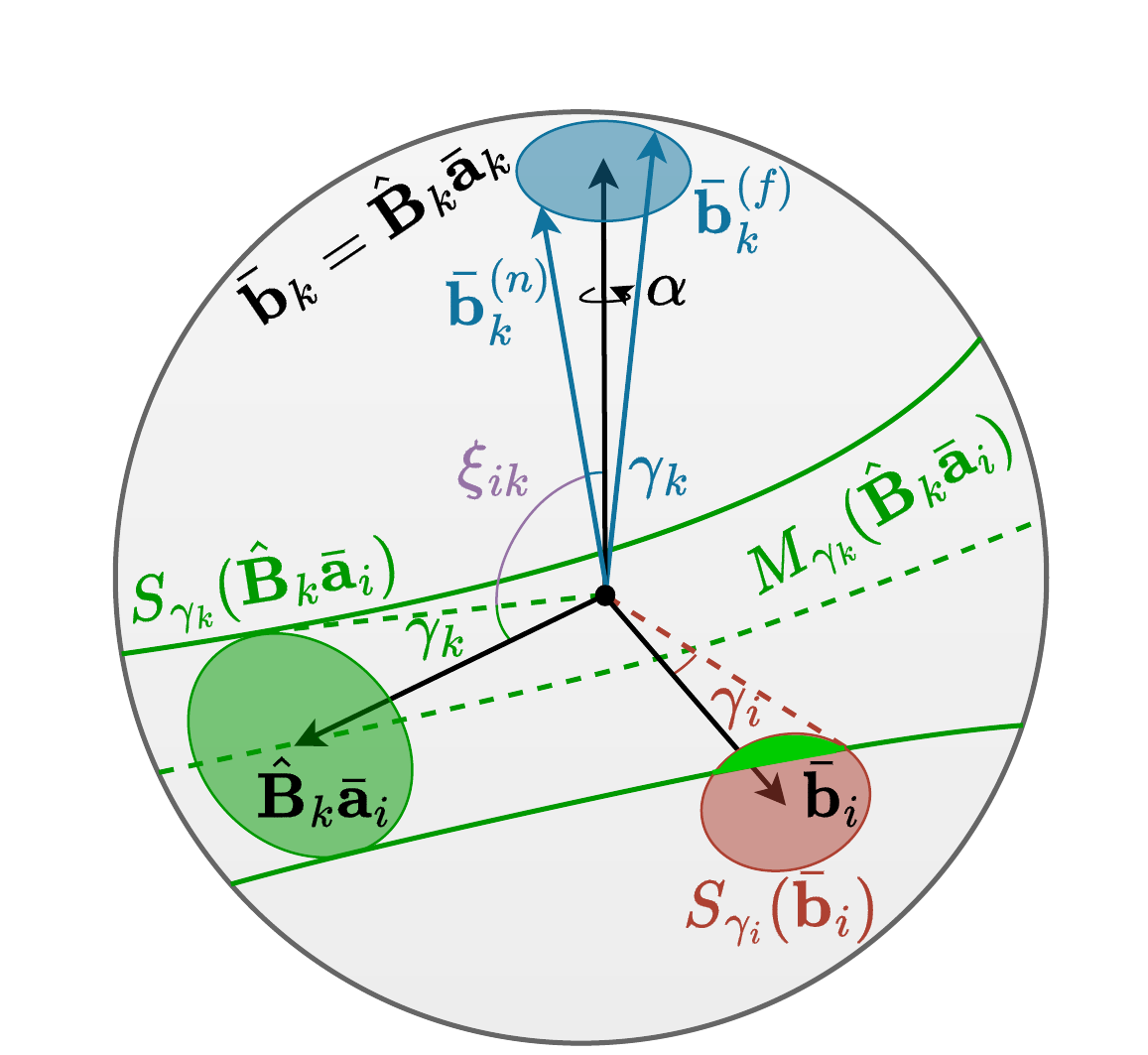}
         \caption{The overlap between $M_{\gamma_i}(\hat{\mathbf{B}}_k \Bar{\mathbf{a}}_i)$ and $S_{\gamma_i}(\Bar{\mathbf{b}}_i)$ shows that there exists a rotation to align both $i$ and $k$ and thus $\mathbb{I}_R = 1$.}
         \label{fig:consensus_func:rotation}
     \end{subfigure}
     \hfill
     \begin{subfigure}[b]{0.27\textwidth}
         \centering
         \includegraphics[width=.8\textwidth]{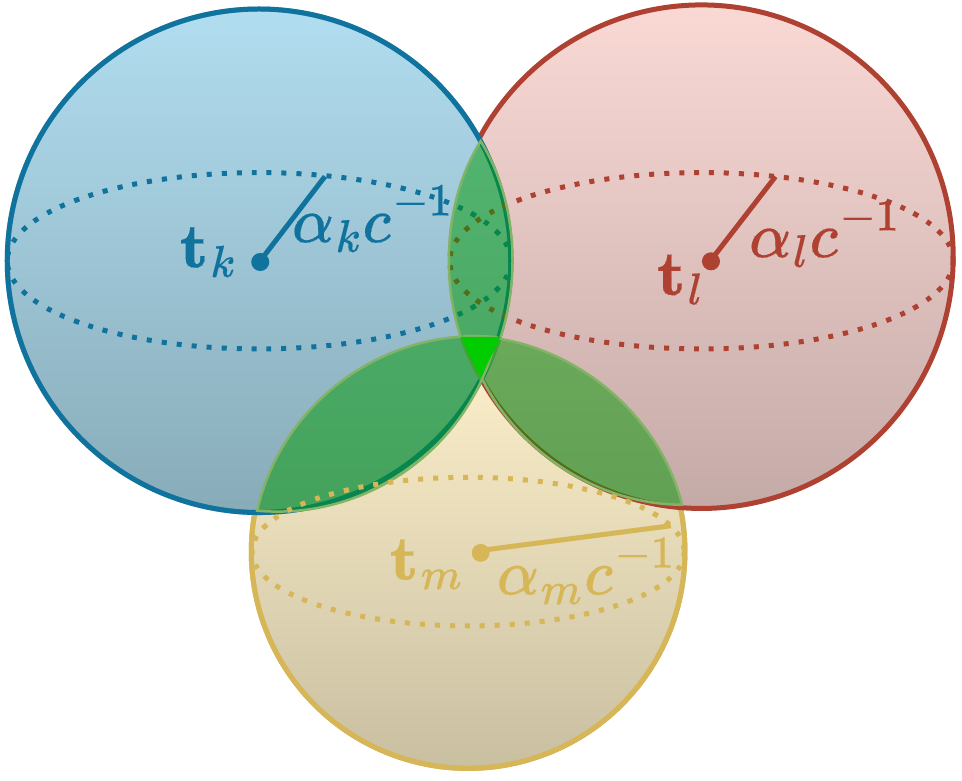}
         \caption{Areas around translation measurements $\mathbf{t}_k$, $\mathbf{t}_m$ and $\mathbf{t}_l$. $\mathbb{I}_t = 1$ for all three pairs $(\mathbf{t}_l, \mathbf{t}_m)$, $(\mathbf{t}_l, \mathbf{t}_k)$ and $(\mathbf{t}_m, \mathbf{t}_k)$}
     \end{subfigure}
\caption{Visualization of the topology of our proposed consensus functions. Dark green shows pairwise consensus and bright green the maximum consensus.}
\label{fig:cons_functs}
\vspace*{-5mm}
\end{figure*}

First, we decouple the overall MC objective for point cloud registration into scale, rotation, and translation estimation. We derive our proposed consensus functions for each objective to map the inlier structure into a graph representation (see figure~\ref{fig:cons_functs}). Using those consensus functions, we introduce our robust registration method \textit{GMCR} and provide implementation details.
%%%%%%%%%%%%%%%%%%%%%%%%%%%%%%%%%%%%%%%%%%%%%%%%%%%%%%%%%%%%%%%%%%%%%%%%%%%%%%%

%%%%%%%%%%%%%%%%%%%%%%%%%%%%%%%%%%%%%%%%%%%%%%%%%%%%%%%%%%%%%%%%%%%%%%%%%%%%%%%

\subsection{Problem Definition}

In correspondence-based rigid point cloud registration, points from set $\mathbf{a}_i \in \mathcal{A}$ are related to points $\mathbf{b}_i \in \mathcal{B}$ by some scaling factor $s\in \mathbb{R}$, rotation $\mathbf{R} \in SO(3)$ and translation $\mathbf{t} \in \mathbb{R}^3$. A set of point-to-point correspondences in the form $\mathcal{C} = \{(\mathbf{a}_i, \mathbf{b}_i): \mathbf{a}_i \in \mathcal{A}, \mathbf{b}_i \in \mathcal{B} \}_{i=1}^{N_\mathcal{C}}$ is given. Due to faulty data association, $\mathcal{C}$ contains an unknown amount of outliers. This can be summarized in the following model:
\begin{equation}
    \mathbf{b}_i = s \mathbf{R} \mathbf{a}_i + \mathbf{t} + \mathbf{o}_i + \boldsymbol{\epsilon}_i \quad .
    \label{eq:gen_model_pcr}
\end{equation}
Therein $\mathbf{o}_i = \mathbf{0}$ iff $i$ is an inlier and $\boldsymbol{\epsilon}_i$ resembles noise. To estimate the parameters robustly we formulate a \textit{Maximum Consensus} objective with an adaptive noise bound $\beta_i$ and a quadratic residual as
\begin{align}
    &\max_{s, \mathbf{R}, \mathbf{t}} \sum_{i=1}^{N_\mathcal{C}} \mathbb{I} (\frac{1}{\beta_i^2} || \mathbf{b}_i - s \mathbf{R} \mathbf{a}_i - \mathbf{t} ||^2 \leq c^{-2}) \quad .
    \label{eq:pcr_maxcon}
\end{align}
$c^{-2}$ is the inlier threshold and $\mathbb{I}$ evaluates to $1$ if the expression in the bracket is true and else to $0$. Furthermore, for a given estimate we define the consensus set as the correspondences for which the residual is below the threshold as
\begin{equation}
    \mathcal{I}(s, \mathbf{R}, \mathbf{t}) = \{i: (\frac{1}{\beta_i^2} || \mathbf{b}_i - s \mathbf{R} \mathbf{a}_i - \mathbf{t} ||^2 \leq c^{-2}) \} \quad . \label{eq:consensus_set}
\end{equation}
 
% Finally, we show how to integrate this into a robust point cloud registration framework. 

%\begin{figure*}[t!]
%     \centering
%     \includegraphics[width=.8\textwidth]{RAL-IROS_2022/figures/3_method/fitting_graph_overview_new.png}
%        \caption{Proposed representation \textit{Fitting Graphs} for consensus structure within a set of measurements. In the model space regions are determined for which two measurements can be in a consensus set. Consensus functions can verify the overlap of those regions and results in a graph. The maximum clique then gives the inlier set, with which the estimate can be computed in closed form.}
%        \label{fig:three graphs}
%\end{figure*}

\subsection{Decoupling the objective}
The concept of decoupling to simplify the objective was introduced by Arun \textit{et al.}~\cite{Arun1987}. We use the formalism of invariants introduced by Shi \textit{et al.}~\cite{Shi2020} for rotation and translation decoupling.

\subsubsection{Translation invariant}
Using vectors between pairs $i$ and $j$ the invariant is given as $f_t(\mathbf{a}_j, \mathbf{a}_i) = \mathbf{a}_j - \mathbf{a}_i = \Bar{\mathbf{a}}_{ji}$, which is then similarly applied to $\mathcal{B}$. Note that only combining two inliers leads to an inlier pair. Thus the most robust approach is taking all-to-all pairs. The underlying model from Eq.~\eqref{eq:gen_model_pcr} can then be simplified as
\begin{align}
    \Bar{\mathbf{b}}_{ji} &= s\mathbf{R}\Bar{\mathbf{a}}_{ji} + \mathbf{o}_{ji} + \boldsymbol{\epsilon}_{ji} \quad . \label{eq:corresp_est:tims}
\end{align}
The new noise bound $||\boldsymbol{\epsilon}_{ji}|| \leq ||\boldsymbol{\epsilon}_j|| + ||\boldsymbol{\epsilon}_i|| = \beta_j + \beta_i = \delta_{ji} = \delta_k$ is set using the triangle inequality and $\mathbf{o}_{ji}$ is again defined as arbitrary, if either $i$ or $j$ is an outlier. Note that in the objectives $ji = k$ for simplicity. The translation invariant MC-objective is then given by
\begin{equation}
    \max_{s, \mathbf{R}} \sum_{k=1}^{K} \mathbb{I} \left( \frac{1}{\delta^2_k} || \Bar{\mathbf{b}}_k - s\mathbf{R} \Bar{\mathbf{a}}_k ||^2 \leq c^{-2} \right) \quad .
    \label{eq:obj_trans_decoupled}
\end{equation}

%\footnote{Given $\mathbf{x}\in \mathbb{R}^n$, $\mathbf{R}\in SO(n)$:$||\mathbf{R}\mathbf{x}|| = \left( \mathbf{x}^T \mathbf{R}^T \mathbf{R} \mathbf{x}  \right)^2 = ||\mathbf{x}||$ \qed}
\subsubsection{Rotation invariant}
With the rotation invariance of the vector-norm, we define the rotation invariant $f_R(\Bar{\mathbf{a}}_{k}) = || \Bar{\mathbf{a}}_{k} ||$, which is again also applied to $\Bar{\mathbf{b}}_{k}$. With this invariant we obtain
\begin{equation}
     || \Bar{\mathbf{b}}_{k} || = || s\mathbf{R}\Bar{\mathbf{a}}_{k} + \mathbf{o}_{k} + \boldsymbol{\epsilon}_{k} || \quad .
\end{equation}
To set the noise bound we again use the triangle inequality
\begin{equation}
    || s\mathbf{R}\Bar{\mathbf{a}}_{k} || - \delta_k \leq || s\mathbf{R}\Bar{\mathbf{a}}_{k} + \boldsymbol{\epsilon}_{k} || \leq || s\mathbf{R}\Bar{\mathbf{a}}_{k} || + \delta_k
\end{equation}
and define $\alpha_k = \frac{\delta_k}{|| \Bar{\mathbf{a}}_k ||}$ to arrive at 
\begin{align}
    ||\Bar{\mathbf{b}}_k || &= s ||\Bar{\mathbf{a}}_{k} || + \Tilde{o}_k + \Tilde{\epsilon}_k \\
    s_k &= s + \Tilde{o}_k + \Tilde{\epsilon}_k \quad .
    \label{eq:corresp_est:trims}
\end{align}
$\Tilde{o}_k$ is defined as $\mathbf{o}_{k}$ previously and $s_k = || \Bar{\mathbf{b}}_k || / || \Bar{\mathbf{a}}_k ||$. Thus the objective, which only depends on scale $s$, is given by
\begin{equation}
    \max_{s} \sum_{k=1}^{K} \mathbb{I} \left( \frac{1}{\alpha^2_k} (s-s_k)^2 \leq c^{-2} \right) \quad .
    \label{eq:corresp_est:scale_obj}
\end{equation}

\subsection{Graph-based Maximum Consensus Estimation}\label{subsec:graph_based_mc}
To solve the objectives Eq. \eqref{eq:corresp_est:scale_obj}, \eqref{eq:obj_trans_decoupled}, and \eqref{eq:pcr_maxcon} we propose our graph-based approach for maximum consensus estimation. To this end, we introduce the concept of consensus functions, which map the consensus structure of measurements w.r.t. the MC-objective to a graph representation. The model to fit to data $\mathcal{D}$ containing measurements $(\mathbf{x}_i, \mathbf{y}_i)$ is given by $\mathbb{M}$. With the consensus set $\mathcal{I}_m$ we define the consensus function $\mathbb{I}_\mathbf{m}$ as: 
%In the general case, we seek to fit a model $\mathbb{M}$ to a dataset $\mathcal{D}$ with measurements $(\mathbf{x}_i, \mathbf{y}_i)$ using maximum consensus estimation. We denote the consensus set as $\mathcal{I}_\mathbf{m}$. $\mathcal{D}$ contains an unknown amount of outliers. For $\mathbb{M}$ and the residual we construct a consensus function $\mathbb{I}_\mathbf{m}$ according to the following definition:
\begin{proposition} \label{proposition:consens_function}
Let there be a function $\mathbb{I}_\mathbf{m}$ that can verify pairwise if measurements are in at least one consensus set together. Formally $\mathbb{I}_\mathbf{m} (\mathbf{x}_i, \mathbf{y}_i, \mathbf{x}_j, \mathbf{y}_j) = 1$ iff $\exists \mathbf{m} \text{ such that } i,j \in \mathcal{I}_\mathbf{m}(\mathbf{m})$ and $0$ else.
\end{proposition}
\noindent
The consensus function is then applied to all pairs to construct the graph $\mathcal{G}_\mathbf{m}$, which encodes the consensus structure:
\begin{proposition} \label{proposition:fitting_graphs}
We define an undirected graph $\mathcal{G}_\mathbf{m} = \{ \mathcal{V}, \mathcal{E} \}$ with vertex set $\mathcal{V}$ and edge set $\mathcal{E}$. Each vertex corresponds to one measurement $\mathbf{x}_i, \mathbf{y}_i \in \mathbb{D}$, therefore the graph has $N$ vertices. Edges are present if $\exists \mathbf{m} $ such that $i,j \in \mathcal{I}_\mathbf{m}(\mathbf{m})$.
\end{proposition}
\noindent
Intuitively one measurement $(\mathbf{x}_i, \mathbf{y}_i)$ corresponds to one $\mathbf{m}_i$. Given $\mathcal{G}_\mathbf{m}$, the maximum consensus set is given by nodes in the maximum clique:
\begin{theorem} \label{theorem:cliques_fitting_graphs}
    Given a graph $\mathcal{G}_\mathbf{m}$ according to definition~\ref{proposition:fitting_graphs}, which was constructed using the consensus functions from definition~\ref{proposition:consens_function}, then the nodes of the maximum clique provide a tight approximation of the maximum consensus set $\mathcal{I}^* _\mathbf{m}$ of the underlying objective. 
\end{theorem}
The tightness of the representation to the actual consensus structure depends on how well pairwise consensus in $\mathbb{I}_\mathbf{m}$ does relate to actual consensus. We show empirically that the representation is tight, but leave a formal proof for future work. In the following, we will detail the concept for scale, rotation, and translation estimation.
\begin{figure*}[t!]
    \centering
    \includegraphics[width=0.8\textwidth]{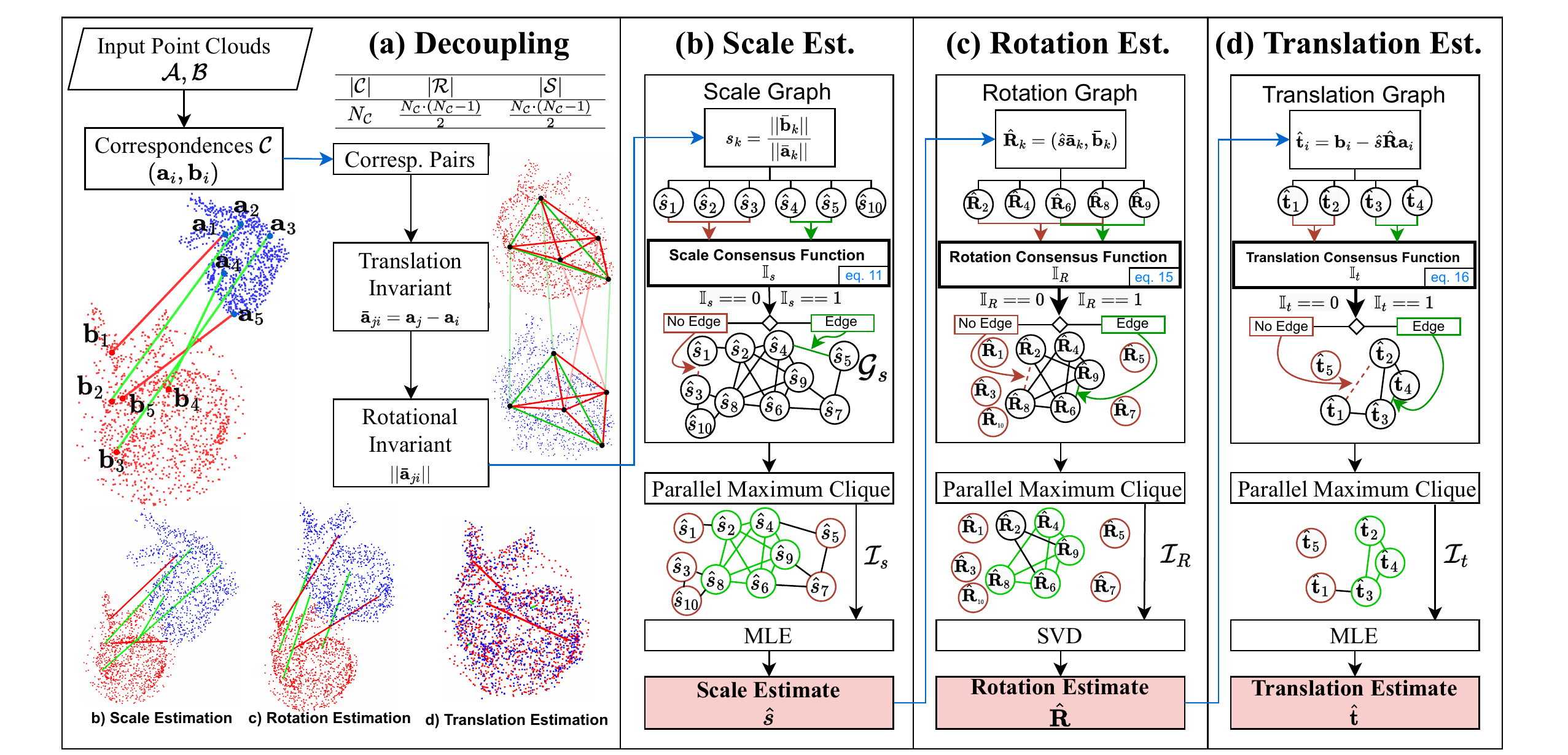}
    \caption{Proposed robust registration framework \textit{GMCR}. Given a set of point-to-point correspondences, we decouple the objective using invariants. We then successively build graphs for the MC-objectives and find the maximum consensus set as the maximum clique. The final estimate is given in closed-form.}
%    Invariants are used to decouple the objective into scale, rotation, and translation estimation. For each objective we construct \textit{Fitting Graphs} using the proposed consensus functions, find the inlier set with the maximum clique and compute the estimate in closed form.}
    \label{fig:framwork}
    \vspace*{-5mm}
\end{figure*}

\subsubsection{Scale Estimation}
For scale estimation (see Eq.~\eqref{eq:corresp_est:scale_obj}) we notice that there exists an interval $[s_k - \alpha_k c, s_k + \alpha_k c]$ around each scale measurement $s_k$ in $s$, for which $k$ is considered as inlier. From two overlapping intervals of $s_k$ and $s_j$, we can deduce that $k$ and $j$ are in at least one consensus set together. Thus using definition~\ref{proposition:consens_function} we propose the scale consensus function as (see fig.~\ref{fig:cons_functs} (a)) 

\begin{equation}
    \mathbb{I}_s(s_k, s_j) = 
    \begin{cases} 
    1  &\text{if } (s_k-\alpha_k c \leq s_j + \alpha_j c) \wedge \;\;\\
    & (s_j - \alpha_j c \leq s_k + \alpha_k c)\\
    0 &\text{\textit{O.T.}}
    \end{cases} \quad .
    \label{eq:consensus:scale}
\end{equation}

\subsubsection{Rotation Estimation}
For rotation estimation (see Eq.~\eqref{eq:obj_trans_decoupled}) we follow the same idea as in scale estimation. The derivation of the consensus function is inspired by GORE~\cite{Bustos2018}. First, we assume that $k$ is an actual inlier. Then there exists a rotation $\mathbf{R}_k$ such that 

\begin{align}
    || \Bar{\mathbf{b}}_k - \hat{s}\mathbf{R}_k \Bar{\mathbf{a}}_k ||^2 &\leq \frac{\delta_k^2}{c^2} \\
    \angle(\Bar{\mathbf{b}}_k, \mathbf{R}_k \Bar{\mathbf{a}}_k') &\leq \gamma_k \quad .
    \label{eq:rotation_consensus_1}
\end{align}

Note that the angular error $\gamma_k$ is defined such that the same set of rotations is admitted. We denote the rotation that perfectly aligns $\Bar{\mathbf{b}}_k$ and $\Bar{\mathbf{a}}_k$ as $\hat{\mathbf{B}}_k$. Since a rotation defined by two vectors is not unique, the overall rotation $\mathbf{R}_k$ is composed of an additional rotation $\mathbf{A}_k$ around axis $\mathbf{B}_k \Bar{\mathbf{a}}_k$ as $\mathbf{R}_k = \mathbf{A}_k \mathbf{B}_k$. Formally the region around any measurement $\mathbf{x}$ bounded by angular deviation $\epsilon$ is given by $S_{\epsilon} (\mathbf{x})$\footnote{$S_{\epsilon} (\mathbf{x}) = \{ \mathbf{y} \in \mathbb{R}^3 | \angle(\mathbf{y}, \mathbf{x}) \leq \epsilon, || \mathbf{x} || = ||\mathbf{y}||\}$}. For now let us assume, that $\mathbf{A}_k = \mathbf{I}_3$. Then the region, in which any other measurement $i \neq k$ can lie, is given by $S_{\gamma_k}(\hat{B}_k \Bar{\mathbf{a_i}})$. For the measurement $i$ to be a possible inlier, this region has to overlap with $S_{\gamma_i}(\Bar{\mathbf{b_i}})$, i.e. $S_{\gamma_k}(\hat{B}_k \Bar{\mathbf{a_i}}) \cap S_{\gamma_i}(\Bar{\mathbf{b_i}}) \neq \emptyset$. In other words, there has to exist a rotation, that rotates $i$ and $k$ into their inlier region. For unknown $\mathbf{A}_k$ we define
\begin{equation}
    M_{k}(\mathbf{x}) = \{\mathbf{y} \in \mathbb{R}^3 | \mathbf{y} \in S_{\gamma_k} (\mathbf{A}\mathbf{x}), \mathbf{A} \in \mathcal{Z}_z, \mathbf{z} \in S_{\gamma_k} (\Bar{\mathbf{b}}_k)\}
\end{equation}
as the possible final positions of $\Bar{\mathbf{a}}_i$, where the set of rotations around an axis $\mathbf{z}$ is given by $\mathcal{Z}_z$\footnote{$\mathcal{Z}_z = \{ \exp(\Theta [\mathbf{z}]_\times ]) | \Theta \in [-\pi, \pi] \}$}. With this, we can formulate the consensus function for $i$ and $k$ as (see fig. \ref{fig:consensus_func:rotation} (b))
\begin{equation}
    \mathbb{I}_R(\Bar{\mathbf{a}}_i, \Bar{\mathbf{a}}_k, \Bar{\mathbf{b}}_i, \Bar{\mathbf{b}}_k) = 
    \begin{cases} 
    1 &\text{if } S_{\gamma_i}(\Bar{\mathbf{b}}_i) \cap \mathcal{M}_{k}(\hat{B}_k\Bar{\mathbf{a}}_i) \neq \emptyset \\
    0 &\text{\textit{O.T.}}
    \end{cases} \quad .
    \label{eq:consensus:rotation}
\end{equation}
$M_{k}(\mathbf{x})$ is a region bound by two circles on the sphere (see figure~\ref{fig:consensus_func:rotation}). To simplify the implementation, we use $\xi_{ik}$ as the constant angle between the upper circle bounding $M_{k}(\mathbf{x})$ and its centroid $\Bar{\mathbf{b}}_k^{(n)}$ to check for full overlap and then negate the result.
%To simplify the implementation, instead of defining $M_{k}(\mathbf{x})$, we check for no overlap and negate the result. We notice that  $\xi_{ik}$ is the constant angle between the upper circle bounding $M_{k}(\mathbf{x})$ and its centroid $\Bar{\mathbf{b}}_k^{(n)}$. Full overlap with either the upper or lower circle can then be checked using the angular deviation. A graphical depiction is given in figure~\ref{fig:cons_functs}(b). 

\subsubsection{Translation Estimation}
We apply the same concept as in scale and rotation to translation estimation (see Eq.~\eqref{eq:obj_trans_decoupled}). The area of consensus for $\mathbf{t}_i$ is given by a 3D sphere with the center $\mathbf{t}_i = \mathbf{b}_i - \hat{s} \hat{\mathbf{R}} \mathbf{a}_i$ and the radius $\beta_i / c$. Again, there only exists an estimate $\mathbf{t}$ for two measurements $i$ and $j$ to be both considered an inlier, if their spheres overlap. Therefore, we define the translation consensus function as (see fig.~\ref{fig:cons_functs}(c)) 
\raggedbottom
\begin{equation}
    \mathbb{I}_t(\hat{\mathbf{t}}_i, \hat{\mathbf{t}}_j) = \begin{cases}
    1 & \text{if  } ||\hat{\mathbf{t}}_j - \hat{\mathbf{t}}_i|| \leq \frac{\beta_j + \beta_i}{c} \\
    0 & \text{\textit{O.T.}}
    \end{cases} \quad .
    \label{eq:corresp_est:translation_consensus_funct}
\end{equation}
We show a visualization in figure~\ref{fig:cons_functs}(c).
%\begin{figure*}[h!]
%\centering
%%\includegraphics[width=.95\textwidth]{ICRA_2023/results/stanford_plot.png}
%\caption{Registration results for scale, rotation and translation on the %\textit{Stanford Scanning Repository}.}
%\label{fig:result:complete_registration}
%\end{figure*}

\subsection{Graph-based Maximum Consensus Registration (GMCR)}
The proposed graph-based approaches for scale, rotation, and translation estimation are integrated into one robust registration framework (see figure~\ref{fig:framwork}). After extracting $N_\mathcal{C} ( N_\mathcal{C} - 1) / 2$ scale measurements using the translation and rotation invariant, we estimate scale, rotation, and translation successively with our graph-based MC-approach as described in section~\ref{subsec:graph_based_mc}. The implementation of the consensus functions is straightforward since they only involve pairwise comparisons. Furthermore, our approach allows for any off-the-shelf maximum clique solver. Note that although maximum clique computation is NP-hard in general as well~\cite{rossi2013parallel}, we observe low runtimes using the Parallel-Maximum-Clique (PMC) solver~\cite{rossi2013parallel} since the \textit{GMCR} graphs get sparser with increasing outlier rates. 
After finding the maximum clique, we compute the final estimate using a closed-form solution for the respective objective with a no-outlier assumption. In the rotation case, we use \textit{Arun et al.'s} method~\cite{Arun1987}. If scale is already known, we prune outliers using consistency graphs as proposed by \textit{Parra et al.}~\cite{Parra2019a}. Outlier measurements are removed after each estimation step. Our framework is implemented in python using CuPy for constructing the adjacency matrices of the graphs. For PMC we use the provided python bindings. 
\section{Experiments}
First, we evaluate our proposed method \textit{GMCR} on a synthetic registration benchmark. We also perform experiments on a real object localization task for robotic manipulation and on a lidar scan-matching dataset. Furthermore, we evaluate \textit{GMCR} and other robust registration methods in the presence of structured outliers. Rotation errors are reported as angular deviation using the geodesic distance $(| \arccos ( (tr (\hat{\mathbf{R}}^T \mathbf{R}^o ) - 1 ) / 2 )| )$ and translation errors are given as $|| \mathbf{t}^o - \hat{\mathbf{t}} ||$. Combined errors are reported as ADD error $||\hat{\mathbf{b}}_i - \mathbf{b}^o_i||$~\cite{hodavn2016evaluation}, where $\hat{\mathbf{b}}_i$ are the estimated points and $\mathbf{b}^o_i$ the points transformed with the ground truth.

%\footnote{Implementation \href{http://www.open3d.org/}{http://www.open3d.org/}}
%\footnote{Implementation: \href{https://speciale.ar/}{https://speciale.ar/}}
\subsection{Synthetic Data}\label{exp:mc_synth}
\begin{figure}[t!]
    \centering
    \includegraphics[width=\linewidth]{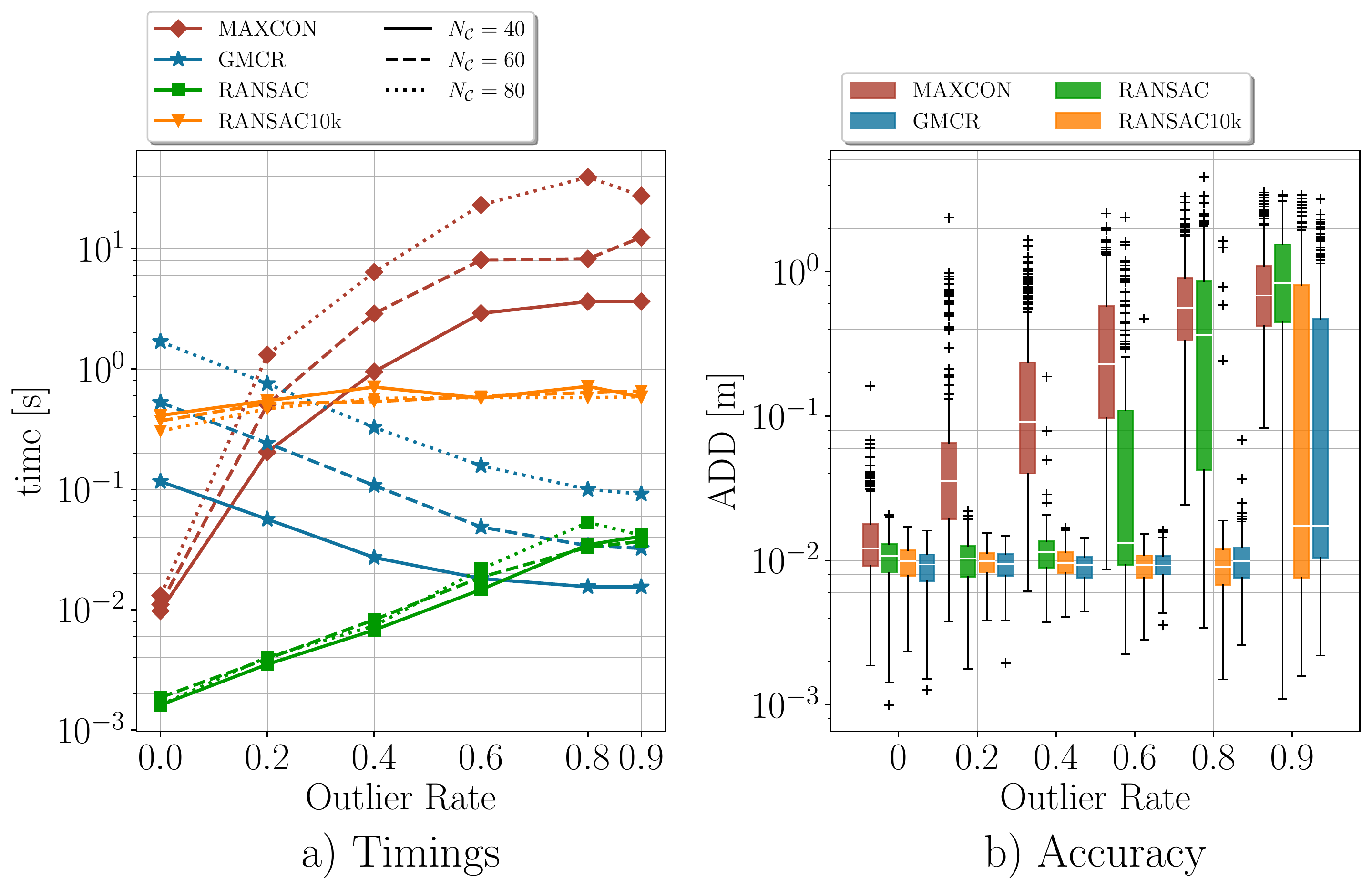}
    \caption{Time and accuracy results on Stanford Scanning Repository.}
    \label{fig:maxcon_comparison}
\end{figure}

We use 7 point cloud models from the Stanford Scanning Repository namely \textit{Armadillo}, \textit{Buddha}, \textit{Bunny}, \textit{Dragon}, \textit{Happy Buddha}, \textit{Lucy}, and \textit{Statue} to evaluate \textit{GMCR} and other state-of-the-art maximum consensus registration approaches. For each mesh, we sample $N=1500$ points and re-scale it to a $1m \times 1m \times 1m$ cube. To simulate sensor corruptions in the target point cloud we sample noise uniformly from $[-0.01m, 0.01m]_3$. Furthermore, $200$ points on a sphere centered at the target point cloud are added. We randomly apply a transformation from $s^o \in [1, 5]$, $\mathbf{R}^o \in SO(3)$, and $\mathbf{t}^o \in [-1.5m, 1.5m]_3$ to the target point cloud. We perform the experiment for $N_\mathcal{C}=\{ 40, 60, 80 \}$ correspondences with $10$ Monte-Carlo runs per object and outlier rate. For \textit{GMCR} we use $\beta_i = 0.02$ and $c=1$. As a comparison, we use \textit{RANSAC} with $10000$ iterations and a confidence bound of $0.99$. Furthermore, we use a \textit{RANSAC10k} variant with $10000$ fixed iterations. Also, we evaluated Speciale \textit{et al.}'s method using the provided implementation as is, which we denote as \textit{MAXCON}. We compare timings and the matching error. Results are shown in figure \ref{fig:maxcon_comparison}.
\subsection{Robust Registration with structured outliers}\label{sec:structured_exp}
\begin{figure}[t!]
    \centering
    \includegraphics[width=\linewidth]{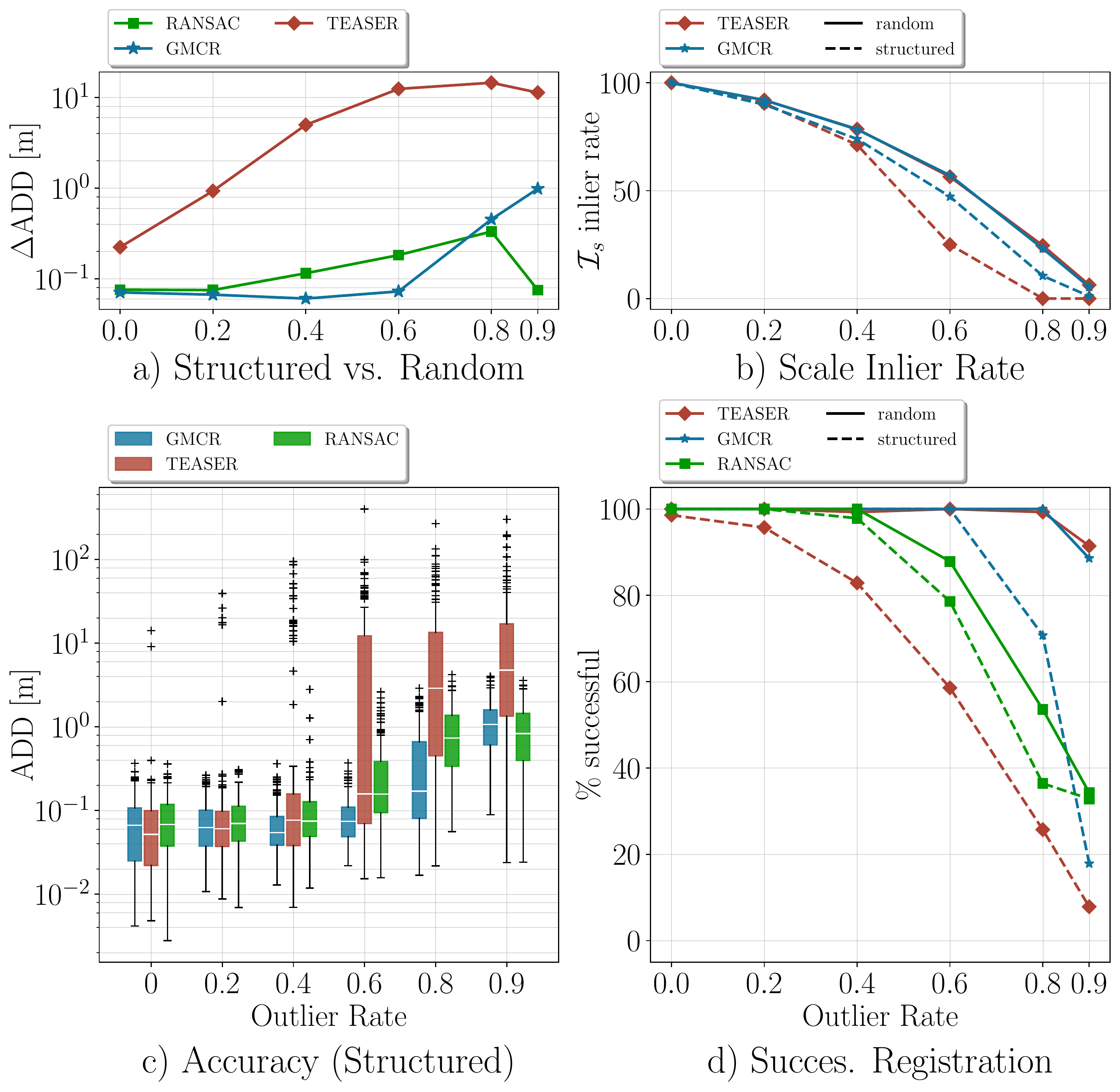}
    \caption{Results on synthetic structured outlier benchmark.}
    \label{fig:structured_outliers}
\end{figure}

To evaluate \textit{GMCR} in the worst-case setting of only structured outliers, we use the same benchmark as in section \ref{exp:mc_synth} with $N_\mathcal{C}=90$. However, inliers are now given by nearest neighbors of the true inliers, which is meant to simulate slight errors in data association. Furthermore, to simulate structured errors in data association, we first sample a set of random bases and then sample the remaining outliers randomly distributed over the base's neighborhoods. We compare against \textit{TEASER++} and \textit{RANSAC}. Both \textit{GMCR} and \textit{TEASER++}'s noise bound is empirically set to $0.07$ and the threshold to $1$. \textit{FGR} does not estimate scale and Speciale \textit{et al.}'s method does not scale for high outlier rates. We report accuracy over varying outlier rates, percent of successfully registered samples (residual $< 0.3m$), average inlier rate in the consensus set for \textit{GMCR} and \textit{TEASER++} and the difference in median for structured outlier sampling and random sampling. Results are shown in figure \ref{fig:structured_outliers}.

\subsection{3D Real Object Localization with Robotics Setup}
Furthermore, we investigate \textit{GMCR} on a real robotic object pose estimation task. We use an Azure Kinect RGB-D sensor attached to the end-effector of a 7-Degree-of-Freedom manipulator robot. Six real-world objects are placed in 5 scenes shuffled in position and orientation. The robot captures 5 different viewpoints as target point clouds per scene. After removing the ground plane, we evaluate three different settings. In the first two experiments, we find inlier correspondences using nearest neighbor search. 50\% Outliers are then sampled according to the strategies defined in sections \ref{exp:mc_synth} and \ref{sec:structured_exp}. Finally, to simulate a real registration pipeline, we use the Fast-Point-Feature-Histogram (FPFH)~\cite{rusu2009fast} approach to estimate point-wise features and then find a set of putative correspondences using nearest neighbor search in feature space. A random transformation sampled from $s^o \in [0.5, 2]$, $\mathbf{R}^o \in SO(3)$, and $\mathbf{t}^o \in [-1m, 1m]_3$ in each viewpoint is applied. Results are reported as \textit{ADD} in table \ref{tab:real_exp}.

% Please add the following required packages to your document preamble:
% \usepackage{booktabs}
\begin{table*}[]
\centering
\caption{Results on real robotic object localization benchmark for random outlier sampling, structured outlier sampling and using FPFH correspondences. Errors are computed as ADD.}
\label{tab:real_exp}
\begin{tabular}{cccccccc}
\hline
                            &             & \shortstack{\includegraphics[width=0.2cm]{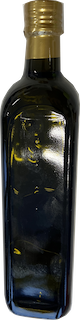} \\ Object 1 [m]}                & \shortstack{\includegraphics[width=0.6cm]{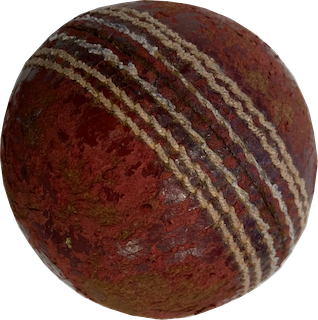} \\ Object 2 [m]}            & \shortstack{\includegraphics[width=0.6cm]{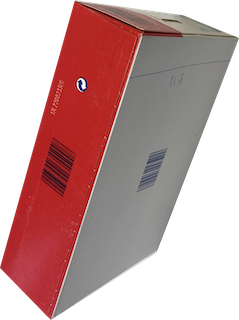} \\ Object 3 [m]}      &    \shortstack{\includegraphics[width=0.45cm]{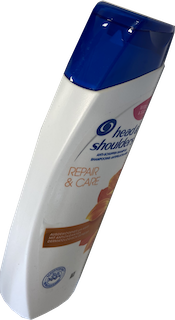} \\ Object 4 [m]}       & \shortstack{\includegraphics[width=0.7cm]{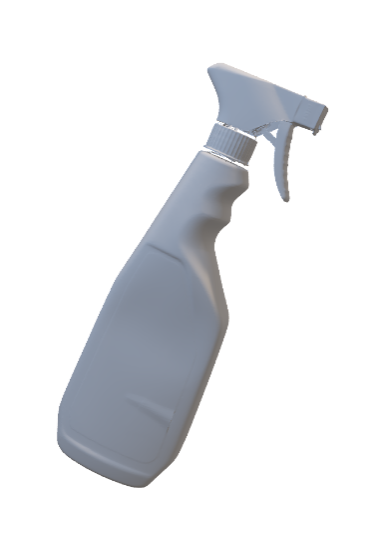} \\ Object 5 [m]}               & \shortstack{\includegraphics[width=0.3cm]{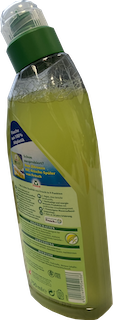} \\ Object 6 [m]}                    \\ \hline
\multirow{3}{*}{Random}     & RANSAC      & $0.216 \pm 0.075$          & $0.077 \pm 0.023$          & $0.109 \pm 0.085$          & $0.187 \pm 0.066$          & $ 0.25 \pm 0.11$           & $0.106 \pm 0.079$          \\
                            & TEASER++    & $\mathbf{0.010 \pm 0.007}$ & $\mathbf{0.013 \pm 0.006}$ & $0.010 \pm 0.005$ & $0.020 \pm 0.008$          & $0.027 \pm 0.013$          & $\mathbf{0.010 \pm 0.006}$ \\
                            & GMCR(Ours)  & $0.012 \pm 0.008$          & $0.015 \pm 0.007$          & $\mathbf{0.010 \pm 0.003}$ & $\mathbf{0.014 \pm 0.008}$ & $\mathbf{0.018 \pm 0.004}$ & $0.012 \pm 0.007$          \\ \hline
\multirow{3}{*}{Structured} & RANSAC      & $0.112 \pm 0.101$          & $0.074 \pm 0.044$          & $0.208 \pm 0.106$          & $0.141 \pm 0.077$          & $0.104 \pm 0.073$          & $0.193 \pm 0.082$          \\
                            & TEASER++    & $0.020 \pm 0.013$          & $0.983 \pm 0.95$           & $1.412 \pm 1.405$          & $0.357 \pm 0.342$          & $\mathbf{0.013 \pm 0.012}$ & $0.386 \pm 0.377$          \\
                            & GMCR (Ours) & $\mathbf{0.015 \pm 0.006}$ & $\mathbf{0.013 \pm 0.006}$ & $\mathbf{0.013 \pm 0.007}$ & $\mathbf{0.024 \pm 0.011}$ & $0.021 \pm 0.011$          & $\mathbf{0.010 \pm 0.005}$ \\ \hline
\multirow{3}{*}{FPFH}       & RANSAC      & $0.293 \pm 0.093$          & $0.097 \pm 0.049$          & $0.291 \pm 0.102$          & $0.288 \pm 0.086$          & $0.369 \pm 0.130$          & $0.325 \pm 0.159$          \\
                            & TEASER++    & $0.187 \pm 0.145$          & $0.156 \pm 0.036$          & $0.317 \pm 0.195$          & $0.290 \pm 0.144$          & $0.383 \pm 0.192$          & $\mathbf{0.152 \pm 0.082}$ \\
                            & GMCR (Ours) & $\mathbf{0.098 \pm 0.056}$ & $\mathbf{0.097 \pm 0.019}$ & $\mathbf{0.197 \pm 0.162}$ & $\mathbf{0.269 \pm 0.141}$ & $\mathbf{0.268 \pm 0.175}$ & $0.172 \pm 0.089$          \\ \hline
\end{tabular}

\end{table*}

\subsection{LiDAR Scan Matching}
\begin{table}[]
\centering
\caption{Error on NuScenes-mini lidar scan matching benchmark.}
\label{tab:lidar}
\begin{adjustbox}{max width=\textwidth}
\begin{tabular}{llll}
\hline
         & \multicolumn{1}{c}{$R [deg] \downarrow$} & \multicolumn{1}{c}{$t [m] \downarrow$} & \multicolumn{1}{c}{\% $\uparrow$} \\ \hline
FGR      & $5.25\pm 5.21$          & $0.21 \pm 0.20$         & $49$                   \\
TEASER++ & $0.24 \pm 0.14$         & $\mathbf{0.03 \pm 0.03}$         & $92$                   \\ \hline
GMCR (Ours)     & $\mathbf{0.24 \pm 0.13}$         & $\mathbf{0.03 \pm 0.03}$         & $\mathbf{95}$                   \\ \hline
\end{tabular}
\end{adjustbox}
\end{table}

%\begin{wraptable}{r}{60mm}
%\begin{tabular}{llll}
%\hline
%         & \multicolumn{1}{c}{$R [deg]$} & \multicolumn{1}{c}{$t [m]$} & \multicolumn{1}{c}{\%} \\ \hline
%FGR      & $5.25\pm 5.21$          & $0.21 \pm 0.20$         & $49$                   \\
%TEASER++ & $0.24 \pm 0.14$         & $\mathbf{0.03 \pm 0.03}$         & $92$                   \\ \hline
%GMCR (Ours)     & $\mathbf{0.24 \pm 0.13}$         & $\mathbf{0.03 \pm 0.03}$         & $\mathbf{95}$                   \\ \hline
%\end{tabular}
%\end{wraptable}
Furthermore, we also validate GMCR on a lidar scan-matching benchmark and compare it against other state-of-the-art robust registration approaches. To this end, we use the nuScenes mini-v1.0 dataset~\cite{9156412}. We transform point clouds into a global coordinate frame using the provided ego pose. First, we sample $N=100$ inlier correspondences from the nearest neighbors with distance $<0.2m$. We sample outlier correspondences by taking $50$ random outliers and distributing $N=300$ correspondences randomly in their neighborhood. A random rotation around the z-axis and translation element-wise sampled from $[-10m, 10m]$ are applied to simulate the movement of the vehicle. Results are reported in table \ref{tab:lidar}.

\section{Discussion}\label{sec:discussion}
In our first set of synthetic registration experiments \textit{GMCR} exhibits much lower computation times than \textit{RANSAC10k} and \textit{MAXCON} over varying outlier rates and number of correspondences (see figure \ref{fig:maxcon_comparison} (a)), while being as accurate as \textit{RANSAC10k}(see figure \ref{fig:maxcon_comparison} (b)). The plain RANSAC variant shows lower computation times, but is less robust and starts to deteriorate at about 60\% outliers. In contrast to the other considered methods, \textit{GMCR}'s runtime decreases as the outlier rate increases, which makes it especially interesting for applications with high outlier rates. This is due to graphs getting sparser with increasing outlier rates and our currently used maximum clique solver PMC becoming more efficient with increasing sparsity. All methods do not exhibit stable performance at 90\% outliers. This can be attributed to the absolute size of the inlier set in the case of $N_\mathcal{C}=40$ leading to high variance in performance since we average over the three experiments.
% 2. Structured noise benchmark + Lidar benchmark
Experiments on our highly challenging structured noise benchmark show, that our method is still robust up to a high outlier rate, whereas \textit{RANSAC} and \textit{TEASER++} start deteriorating at about 60\% outliers (see figure \ref{fig:structured_outliers} (c)). The drop in performance can also be seen by the drop in the percentage of successful registrations in figure \ref{fig:structured_outliers} (d). The difference in accuracy compared to random outlier sampling in figure \ref{fig:structured_outliers} (a) shows that TEASER++ is the most affected method in our comparison. From figure~\ref{fig:structured_outliers} (b) we see that the average inlier percentage of the scale consensus set, which was used for computing the scale estimate, is decreasing sharply. Also, we can see that \textit{GMCR} performs comparable in lidar scan matching to \textit{TEASER++} (see table \ref{tab:lidar}), which requires only estimation of rotation and translation. Since outliers caused by ambiguities in scale can mostly be removed through consistency graph-based outlier removal, we hypothesize that the drop in accuracy caused by structured outliers mainly arises during scale estimation. 
% 3. Real robotic experiment
Furthermore, we performed a real robotic experiment with a set of 6 household objects, where we estimate scale, rotation, and translation (see table \ref{tab:real_exp}). Random outlier sampling yields highly accurate results over all methods, whereas \textit{TEASER++} and ours perform on par. Structured outlier sampling again introduces a big drop in accuracy for other methods, especially for \textit{TEASER++}. Finally, FPFH-based correspondences lead to a moderate performance in all methods, whereas ours still outperforms \textit{TEASER++} and \textit{RANSAC}. Note that our objects exhibit similar surfaces and symmetry, which makes it difficult for local descriptors to capture differentiating features.
% 4. Drawbacks of our method
Because of the NP-hardness of the maximum clique problem and the exact PMC solver, worst-case instances in runtime for \textit{GMCR} are possible. Furthermore, the problem size is currently limited by the adjacency matrix construction and not the clique solver.

\section{Conclusion}\label{sec:conclusion}
We proposed our novel graph-based approach to maximum consensus point cloud registration \textit{GMCR}. Using our proposed consensus functions we map the decoupled objectives to the graph domain, where the maximum consensus set is found as the maximum clique. \textit{GMCR} compares favorably with state-of-the-art robust registration methods in timing, accuracy, and robustness on synthetic and real benchmarks. 
% Future work includes the investigation of approximate but faster clique solvers and deriving theoretical bounds on the estimation errors.
As future work, we would like to extend our proposed approach for general robust perception tasks including cross-modal perception~\cite{murali2022deep, murali2022towards}, robotic manipulation~\cite{kaboli2016tactile, yao2017tactile, kaboli2016re, feng2018active}, mobile robotics~\cite{murali2022intelligent}, and other sensing modalities~\cite{liu2022neuro, sandamirskaya2022neuromorphic}.

\section*{\footnotesize ACKNOWLEDGMENT}
\footnotesize {We would like to thank Prof. Luca Carlone for the insightful discussions on their work TEASER++.
We would also like to thank Prof. Stephan G\" unnemann for the useful feedback. This work was supported in part by BMW Group to MG, PKM,  MK, and the European Commission via INTUITIVE under Grant 861166, PHASTRAC 101092096  under Grant, iNavigate under Grant 873178 to PKM and MK.}

%%%%%%%%%%%%%%%%%%%%%%%%%%%%%%%%%%%%%%%%%%%%%%%%%%%%%%%%%%%%%%%%%%%%%%%%%%%%%%%%
\newpage
\bibliographystyle{IEEEtran}
\bibliography{root}

\begin{thebibliography}{10}
\providecommand{\url}[1]{#1}
\csname url@rmstyle\endcsname
\providecommand{\newblock}{\relax}
\providecommand{\bibinfo}[2]{#2}
\providecommand\BIBentrySTDinterwordspacing{\spaceskip=0pt\relax}
\providecommand\BIBentryALTinterwordstretchfactor{4}
\providecommand\BIBentryALTinterwordspacing{\spaceskip=\fontdimen2\font plus
\BIBentryALTinterwordstretchfactor\fontdimen3\font minus
  \fontdimen4\font\relax}
\providecommand\BIBforeignlanguage[2]{{%
\expandafter\ifx\csname l@#1\endcsname\relax
\typeout{** WARNING: IEEEtran.bst: No hyphenation pattern has been}%
\typeout{** loaded for the language `#1'. Using the pattern for}%
\typeout{** the default language instead.}%
\else
\language=\csname l@#1\endcsname
\fi
#2}}

\bibitem{bosse2016robust}
M.~Bosse \emph{et~al.}, ``Robust estimation and applications in robotics,''
  \emph{Found. and Trends in Rob.}, vol.~4, no.~4, pp. 225--269, 2016.

\bibitem{murali2022active}
P.~K. Murali, A.~Dutta, M.~Gentner, E.~Burdet, R.~Dahiya, and M.~Kaboli,
  ``Active visuo-tactile interactive robotic perception for accurate object
  pose estimation in dense clutter,'' \emph{IEEE Robotics and Automation
  Letters}, vol.~7, no.~2, pp. 4686--4693, 2022.

\bibitem{pomerleau2013comparing}
F.~Pomerleau, F.~Colas, R.~Siegwart, and S.~Magnenat, ``{Comparing ICP variants
  on real-world data sets: Open-source library and experimental protocol},''
  \emph{Autonomous Robots}, vol.~34, no.~3, pp. 133--148, 2013.

\bibitem{delmerico2018comparison}
J.~Delmerico \emph{et~al.}, ``A comparison of volumetric information gain
  metrics for active 3d object reconstruction,'' \emph{Autonomous Robots},
  vol.~42, no.~2, pp. 197--208, 2018.

\bibitem{magnusson2015beyond}
M.~Magnusson \emph{et~al.}, ``Beyond points: Evaluating recent 3d scan-matching
  algorithms,'' in \emph{2015 IEEE Int. Conf. on Rob. and Aut. (ICRA)}.\hskip
  1em plus 0.5em minus 0.4em\relax IEEE, 2015, pp. 3631--3637.

\bibitem{audette2000algorithmic}
M.~A. Audette \emph{et~al.}, ``An algorithmic overview of surface registration
  techniques for medical imaging,'' \emph{Med. image analysis}, vol.~4, no.~3,
  pp. 201--217, 2000.

\bibitem{li2020review}
Q.~Li, O.~Kroemer, Z.~Su, F.~F. Veiga, M.~Kaboli, and H.~J. Ritter, ``A review
  of tactile information: Perception and action through touch,'' \emph{IEEE
  Transactions on Robotics}, vol.~36, no.~6, pp. 1619--1634, 2020.

\bibitem{murali2021active}
P.~K. Murali, M.~Gentner, and M.~Kaboli, ``Active visuo-tactile point cloud
  registration for accurate pose estimation of objects in an unknown
  workspace,'' in \emph{2021 IEEE/RSJ International Conference on Intelligent
  Robots and Systems (IROS)}.\hskip 1em plus 0.5em minus 0.4em\relax IEEE,
  2021, pp. 2838--2844.

\bibitem{murali2022intelligent}
P.~K. Murali, M.~Kaboli, and R.~Dahiya, ``Intelligent in-vehicle interaction
  technologies,'' \emph{Advanced Intelligent Systems}, vol.~4, no.~2, p.
  2100122, 2022.

\bibitem{kaboli2019tactile}
M.~Kaboli, K.~Yao, D.~Feng, and G.~Cheng, ``Tactile-based active object
  discrimination and target object search in an unknown workspace,''
  \emph{Autonomous Robots}, vol.~43, pp. 123--152, 2019.

\bibitem{kaboli2018active}
M.~Kaboli, D.~Feng, and G.~Cheng, ``Active tactile transfer learning for object
  discrimination in an unstructured environment using multimodal robotic
  skin,'' \emph{International Journal of Humanoid Robotics}, vol.~15, no.~01,
  p. 1850001, 2018.

\bibitem{kaboli2017tactile}
M.~Kaboli, D.~Feng, K.~Yao, P.~Lanillos, and G.~Cheng, ``A tactile-based
  framework for active object learning and discrimination using multimodal
  robotic skin,'' \emph{IEEE Robotics and Automation Letters}, vol.~2, no.~4,
  pp. 2143--2150, 2017.

\bibitem{kaboli2018robust}
M.~Kaboli and G.~Cheng, ``Robust tactile descriptors for discriminating objects
  from textural properties via artificial robotic skin,'' \emph{IEEE
  Transactions on Robotics}, vol.~34, no.~4, pp. 985--1003, 2018.

\bibitem{liu2022neuro}
F.~Liu, S.~Deswal, A.~Christou, Y.~Sandamirskaya, M.~Kaboli, and R.~Dahiya,
  ``Neuro-inspired electronic skin for robots,'' \emph{Science Robotics},
  vol.~7, no.~67, p. eabl7344, 2022.

\bibitem{Bustos2018}
A.~P. Bustos and T.~J. Chin, ``{Guaranteed Outlier Removal for Point Cloud
  Registration with Correspondences},'' \emph{IEEE Trans. on Pat. Ana. and
  Mach. Intell.}, vol.~40, no.~12, pp. 2868--2882, 2018.

\bibitem{Parra2019a}
{\'{A}}.~Parra, T.-J. Chin, F.~Neumann, T.~Friedrich, and M.~Katzmann, ``{A
  Practical Maximum Clique Algorithm for Matching with Pairwise Constraints},''
  feb 2019.

\bibitem{guerrero2018pcpnet}
P.~Guerrero, Y.~Kleiman, M.~Ovsjanikov, and N.~J. Mitra, ``Pcpnet learning
  local shape properties from raw point clouds,'' in \emph{Computer Graphics
  Forum}, vol.~37, no.~2.\hskip 1em plus 0.5em minus 0.4em\relax Wiley Online
  Library, 2018, pp. 75--85.

\bibitem{jalobeanu2014unknown}
A.~Jalobeanu and G.~Gon{\c{c}}alves, ``The unknown spatial quality of dense
  point clouds derived from stereo images,'' \emph{IEEE Geoscience and Remote
  Sensing Letters}, vol.~12, no.~5, pp. 1013--1017, 2014.

\bibitem{4458156}
A.~Kanemura, S.-i. Maeda, and S.~Ishii, ``Image superresolution under spatially
  structured noise,'' in \emph{2007 IEEE International Symposium on Signal
  Processing and Information Technology}, 2007, pp. 275--280.

\bibitem{10.5555/993884}
G.~Medioni and S.~B. Kang, \emph{Emerging Topics in Computer Vision}.\hskip 1em
  plus 0.5em minus 0.4em\relax USA: Prentice Hall PTR, 2004.

\bibitem{carlone2022estimation}
L.~Carlone, ``Estimation contracts for outlier-robust geometric perception,''
  \emph{arXiv preprint arXiv:2208.10521}, 2022.

\bibitem{Chin}
T.~J. Chin, Z.~Cai, and F.~Neumann, ``{Robust Fitting in Computer Vision: Easy
  or Hard?}'' \emph{International Journal of Computer Vision}, vol. 128, no.~3,
  pp. 575--587, 2020.

\bibitem{fischler1981random}
M.~A. Fischler and R.~C. Bolles, ``{Random sample consensus: A Paradigm for
  Model Fitting with Applications to Image Analysis and Automated
  Cartography},'' \emph{Com. of the ACM}, vol.~24, no.~6, pp. 381--395, 1981.

\bibitem{10.1007/3-540-47969-4_6}
B.~Tordoff and D.~W. Murray, ``Guided sampling and consensus for motion
  estimation,'' in \emph{Computer Vision --- ECCV 2002}, A.~Heyden, G.~Sparr,
  M.~Nielsen, and P.~Johansen, Eds.\hskip 1em plus 0.5em minus 0.4em\relax
  Berlin, Heidelberg: Springer Berlin Heidelberg, 2002, pp. 82--96.

\bibitem{1467271}
O.~Chum and J.~Matas, ``Matching with prosac - progressive sample consensus,''
  in \emph{2005 IEEE Computer Society Conference on Computer Vision and Pattern
  Recognition (CVPR'05)}, vol.~1, 2005, pp. 220--226 vol. 1.

\bibitem{torr2002bayesian}
P.~H.~S. Torr, ``Bayesian model estimation and selection for epipolar geometry
  and generic manifold fitting,'' \emph{International Journal of Computer
  Vision}, vol.~50, no.~1, pp. 35--61, 2002.

\bibitem{konouchine2005amlesac}
A.~Konouchine, V.~Gaganov, and V.~Veznevets, ``Amlesac: A new maximum
  likelihood robust estimator,'' in \emph{Proc. Graphicon}, vol.~5.\hskip 1em
  plus 0.5em minus 0.4em\relax Citeseer, 2005, pp. 93--100.

\bibitem{sun2021ransic}
L.~Sun, ``Ransic: Fast and highly robust estimation for rotation search and
  point cloud registration using invariant compatibility,'' \emph{IEEE Robotics
  and Automation Letters}, vol.~7, no.~1, pp. 143--150, 2021.

\bibitem{bazin2012globally}
J.-C. Bazin, Y.~Seo, and M.~Pollefeys, ``Globally optimal consensus set
  maximization through rotation search,'' in \emph{Asian Conference on Computer
  Vision}.\hskip 1em plus 0.5em minus 0.4em\relax Springer, 2012, pp. 539--551.

\bibitem{zheng2011deterministically}
Y.~Zheng, S.~Sugimoto, and M.~Okutomi, ``Deterministically maximizing feasible
  subsystem for robust model fitting with unit norm constraint,'' in \emph{CVPR
  2011}.\hskip 1em plus 0.5em minus 0.4em\relax IEEE, 2011, pp. 1825--1832.

\bibitem{olsson2008branch}
C.~Olsson, F.~Kahl, and M.~Oskarsson, ``Branch-and-bound methods for euclidean
  registration problems,'' \emph{IEEE Transactions on Pattern Analysis and
  Machine Intelligence}, vol.~31, no.~5, pp. 783--794, 2008.

\bibitem{li2009consensus}
H.~Li, ``Consensus set maximization with guaranteed global optimality for
  robust geometry estimation,'' in \emph{2009 IEEE 12th International
  Conference on Computer Vision}.\hskip 1em plus 0.5em minus 0.4em\relax IEEE,
  2009, pp. 1074--1080.

\bibitem{chin2015efficient}
T.-J. Chin, P.~Purkait, A.~Eriksson, and D.~Suter, ``Efficient globally optimal
  consensus maximisation with tree search,'' in \emph{Proceedings of the IEEE
  Conference on Computer Vision and Pattern Recognition}, 2015, pp. 2413--2421.

\bibitem{parra2014fast}
A.~Parra~Bustos, T.-J. Chin, and D.~Suter, ``Fast rotation search with
  stereographic projections for 3d registration,'' in \emph{Proceedings of the
  IEEE conference on computer vision and pattern recognition}, 2014, pp.
  3930--3937.

\bibitem{Speciale2017}
P.~Speciale, D.~P. Paudel, M.~R. Oswald, T.~Kroeger, L.~V. Gool, and
  M.~Pollefeys, ``{Consensus maximization with linear matrix inequality
  constraints},'' in \emph{Proceedings - 30th IEEE Conference on Computer
  Vision and Pattern Recognition, CVPR 2017}, vol. 2017-Janua, 2017, pp.
  5048--5056.

\bibitem{Yang2020}
H.~Yang, P.~Antonante, V.~Tzoumas, and L.~Carlone, ``{Graduated Non-Convexity
  for Robust Spatial Perception: From Non-Minimal Solvers to Global Outlier
  Rejection},'' \emph{IEEE Robotics and Automation Letters}, vol.~5, no.~2, pp.
  1127--1134, 2020.

\bibitem{Yang2021}
H.~Yang \emph{et~al.}, ``{Teaser: Fast and certifiable point cloud
  registration},'' \emph{IEEE Trans. on Rob.}, vol.~37, no.~2, pp. 314--333,
  2021.

\bibitem{Zhou2016}
Q.-Y. Zhou \emph{et~al.}, ``Fast global registration,'' in \emph{Euro. conf. on
  comp. vis.}\hskip 1em plus 0.5em minus 0.4em\relax Springer, 2016, pp.
  766--782.

\bibitem{Black1996}
M.~J. Black and A.~Rangarajan, ``{On the unification of line processes, outlier
  rejection, and robust statistics with applications in early vision},''
  \emph{International Journal of Computer Vision}, vol.~19, no.~1, pp. 57--91,
  1996.

\bibitem{Besl1992}
P.~J. Besl and N.~D. McKay, ``{A Method for Registration of 3-D Shapes},''
  \emph{IEEE Trans. on Pat. Ana. and Mac. Intel.}, vol.~14, no.~2, pp.
  239--256, 1992.

\bibitem{Granger2002}
S.~Granger \emph{et~al.}, ``A fast and robust approach for surface
  registration,'' in \emph{Proc. of the 7th Euro. Conf. on Comp. Vis.-Part IV},
  pp. 418--432.

\bibitem{Kaneko2003}
S.~Kaneko \emph{et~al.}, ``{Robust matching of 3D contours using iterative
  closest point algorithm improved by M-estimation},'' \emph{Pat. Rec.},
  vol.~36, no.~9, pp. 2041--2047, 2003.

\bibitem{Chetverikov2005}
D.~Chetverikov \emph{et~al.}, ``{Robust Euclidean alignment of 3D point sets:
  The trimmed iterative closest point algorithm},'' \emph{Im. and Vis. Comp.},
  vol.~23, no.~3, pp. 299--309, 2005.

\bibitem{Maier-Hein2012}
L.~Maier-Hein \emph{et~al.}, ``{Convergent iterative closest-point algorithm to
  accomodate anisotropic and inhomogenous localization error},'' \emph{IEEE
  Trans. on Pat. Ana. and Mac. Intel.}, vol.~34, no.~8, pp. 1520--1532, 2012.

\bibitem{segal2009generalized}
A.~Segal, D.~Haehnel, and S.~Thrun, ``Generalized-icp.'' in \emph{Robotics:
  science and systems}, vol.~2, no.~4.\hskip 1em plus 0.5em minus 0.4em\relax
  Seattle, WA, 2009, p. 435.

\bibitem{murali2022empirical}
P.~K. Murali, R.~Dahiya, and M.~Kaboli, ``An empirical evaluation of various
  information gain criteria for active tactile action selection for pose
  estimation,'' in \emph{2022 IEEE International Conference on Flexible and
  Printable Sensors and Systems (FLEPS)}.\hskip 1em plus 0.5em minus
  0.4em\relax IEEE, 2022, pp. 1--4.

\bibitem{Yang2016}
J.~Yang, H.~Li, D.~Campbell, and Y.~Jia, ``{Go-ICP: A Globally Optimal Solution
  to 3D ICP Point-Set Registration},'' \emph{IEEE Trans. on Pat. Ana. and Mac.
  Intel.}, vol.~38, no.~11, pp. 2241--2254, 2016.

\bibitem{hartley2009global}
R.~I. Hartley and F.~Kahl, ``Global optimization through rotation space
  search,'' \emph{International Journal of Computer Vision}, vol.~82, no.~1,
  pp. 64--79, 2009.

\bibitem{breuel2003implementation}
T.~M. Breuel, ``Implementation techniques for geometric branch-and-bound
  matching methods,'' \emph{Computer Vision and Image Understanding}, vol.~90,
  no.~3, pp. 258--294, 2003.

\bibitem{guo2020deep}
Y.~Guo \emph{et~al.}, ``Deep learning for 3d point clouds: A survey,''
  \emph{IEEE Trans. on Pat. Ana. and Mac. Intel.}, 2020.

\bibitem{Aoki2019}
Y.~Aoki \emph{et~al.}, ``Pointnetlk: Robust \& efficient point cloud
  registration using pointnet,'' in \emph{Proceedings of the IEEE/CVF Conf. on
  Comp. Vis. and Pat. Rec.}, 2019, pp. 7163--7172.

\bibitem{Wang2019a}
Y.~Wang and J.~Solomon, ``{Deep closest point: Learning representations for
  point cloud registration},'' in \emph{Proceedings of the IEEE Int. Conf. on
  Comp. Vis.}, vol. 2019-Octob, 2019, pp. 3522--3531.

\bibitem{Wang2019b}
Y.~Wang \emph{et~al.}, ``Dynamic graph cnn for learning on point clouds,''
  \emph{Acm Trans. On Graphics (tog)}, vol.~38, no.~5, pp. 1--12, 2019.

\bibitem{wicker2019robustness}
M.~Wicker and M.~Kwiatkowska, ``Robustness of 3d deep learning in an
  adversarial setting,'' in \emph{Proceedings of the IEEE/CVF Conference on
  Computer Vision and Pattern Recognition}, 2019, pp. 11\,767--11\,775.

\bibitem{Enqvist2009}
O.~Enqvist, K.~Josephson, and F.~Kahl, ``{Optimal correspondences from pairwise
  constraints},'' in \emph{Proceedings of the IEEE International Conference on
  Computer Vision}, 2009, pp. 1295--1302.

\bibitem{8460217}
J.~G. Mangelson, D.~Dominic, R.~M. Eustice, and R.~Vasudevan, ``Pairwise
  consistent measurement set maximization for robust multi-robot map merging,''
  in \emph{2018 IEEE International Conference on Robotics and Automation
  (ICRA)}, 2018, pp. 2916--2923.

\bibitem{leordeanu2005spectral}
M.~Leordeanu and M.~Hebert, ``A spectral technique for correspondence problems
  using pairwise constraints,'' 2005.

\bibitem{Lusk2021}
P.~C. Lusk \emph{et~al.}, ``{CLIPPER: A Graph-Theoretic Framework for Robust
  Data Association},'' \emph{IEEE Int. Conf. on Rob. and Aut. (ICRA)}, 2021.

\bibitem{Shi2020}
J.~Shi \emph{et~al.}, ``{ROBIN: a Graph-Theoretic Approach to Reject Outliers
  in Robust Estimation using Invariants},'' \emph{IEEE Int. Conf. on Rob. and
  Aut. (ICRA)}, 2021.

\bibitem{Arun1987}
K.~S. Arun \emph{et~al.}, ``{Least-Squares Fitting of Two 3-D Point Sets},''
  \emph{IEEE Trans. on Pat. Ana. and Mach. Intell.}, vol. PAMI-9, no.~5, pp.
  698--700, 1987.

\bibitem{rossi2013parallel}
R.~A. Rossi, D.~F. Gleich, and A.~H. Gebremedhin, ``Parallel maximum clique
  algorithms with applications to network analysis,'' \emph{SIAM J. Sci.
  Comput.}, vol.~37, 2015.

\bibitem{hodavn2016evaluation}
T.~Hoda{\v{n}}, J.~Matas, and {\v{S}}.~Obdr{\v{z}}{\'a}lek, ``On evaluation of
  6d object pose estimation,'' in \emph{European Conference on Computer
  Vision}.\hskip 1em plus 0.5em minus 0.4em\relax Springer, 2016, pp. 606--619.

\bibitem{rusu2009fast}
R.~B. Rusu \emph{et~al.}, ``{Fast Point Feature Histograms (FPFH) for 3D
  registration},'' in \emph{2009 IEEE int. conf. on rob. and aut.}\hskip 1em
  plus 0.5em minus 0.4em\relax IEEE, 2009, pp. 3212--3217.

\bibitem{9156412}
H.~Caesar, V.~Bankiti, A.~H. Lang, S.~Vora, V.~Liong, Q.~Xu, A.~Krishnan,
  Y.~Pan, G.~Baldan, and O.~Beijbom, ``nuscenes: A multimodal dataset for
  autonomous driving,'' in \emph{2020 IEEE/CVF Conference on Computer Vision
  and Pattern Recognition (CVPR)}.\hskip 1em plus 0.5em minus 0.4em\relax Los
  Alamitos, CA, USA: IEEE Computer Society, jun 2020, pp. 11\,618--11\,628.

\bibitem{murali2022deep}
P.~K. Murali, C.~Wang, D.~Lee, R.~Dahiya, and M.~Kaboli, ``Deep active
  cross-modal visuo-tactile transfer learning for robotic object recognition,''
  \emph{IEEE Robotics and Automation Letters}, vol.~7, no.~4, pp. 9557--9564,
  2022.

\bibitem{murali2022towards}
P.~K. Murali, C.~Wang, R.~Dahiya, and M.~Kaboli, ``Towards robust 3d object
  recognition with dense-to-sparse deep domain adaptation,'' in \emph{2022 IEEE
  International Conference on Flexible and Printable Sensors and Systems
  (FLEPS)}.\hskip 1em plus 0.5em minus 0.4em\relax IEEE, 2022, pp. 1--4.

\bibitem{kaboli2016tactile}
M.~Kaboli, K.~Yao, and G.~Cheng, ``Tactile-based manipulation of deformable
  objects with dynamic center of mass,'' in \emph{2016 IEEE-RAS 16th
  International Conference on Humanoid Robots (Humanoids)}.\hskip 1em plus
  0.5em minus 0.4em\relax IEEE, 2016, pp. 752--757.

\bibitem{yao2017tactile}
K.~Yao, M.~Kaboli, and G.~Cheng, ``Tactile-based object center of mass
  exploration and discrimination,'' in \emph{2017 IEEE-RAS 17th International
  Conference on Humanoid Robotics (Humanoids)}.\hskip 1em plus 0.5em minus
  0.4em\relax IEEE, 2017, pp. 876--881.

\bibitem{kaboli2016re}
M.~Kaboli, R.~Walker, and G.~Cheng, ``Re-using prior tactile experience by
  robotic hands to discriminate in-hand objects via texture properties,'' in
  \emph{2016 IEEE International Conference on Robotics and Automation
  (ICRA)}.\hskip 1em plus 0.5em minus 0.4em\relax IEEE, 2016, pp. 2242--2247.

\bibitem{feng2018active}
D.~Feng, M.~Kaboli, and G.~Cheng, ``Active prior tactile knowledge transfer for
  learning tactual properties of new objects,'' \emph{Sensors}, vol.~18, no.~2,
  p. 634, 2018.

\bibitem{sandamirskaya2022neuromorphic}
Y.~Sandamirskaya, M.~Kaboli, J.~Conradt, and T.~Celikel, ``Neuromorphic
  computing hardware and neural architectures for robotics,'' \emph{Science
  Robotics}, vol.~7, no.~67, p. eabl8419, 2022.

\end{thebibliography}

\end{document}